\documentclass{article} % For LaTeX2e
\usepackage{iclr2019_conference,times}

\usepackage{amsmath,amsfonts,bm}

\newcommand{\newterm}[1]{{\bf #1}}

\def\eqref#1{equation~\ref{#1}}
\def\ceil#1{\lceil #1 \rceil}

\def\1{\bm{1}}

\DeclareMathAlphabet{\mathsfit}{\encodingdefault}{\sfdefault}{m}{sl}
\SetMathAlphabet{\mathsfit}{bold}{\encodingdefault}{\sfdefault}{bx}{n}

\def\emO{{O}}

\def\emW{{W}}
\def\emX{{X}}

\newcommand{\R}{\mathbb{R}}

\newcommand{\softmax}{\mathrm{softmax}}
\newcommand{\sigmoid}{\sigma}

\def\Sdconvfull{Lightweight convolution}
\def\sdconvfull{lightweight convolution}
\def\sdconv{LightConv}
\def\Tvsdconvfull{Dynamic convolution}
\def\tvsdconvfull{dynamic convolution}
\def\tvsdconv{DynamicConv}

\def\gbw{Billion word}

\def\cnndm{CNN-DailyMail}
\def\iwslt{IWSLT}
\def\ende{WMT En-De}
\def\enfr{WMT En-Fr}
\def\zhen{WMT Zh-En}

\usepackage{hyperref}
\usepackage{url}
\usepackage{booktabs}
\usepackage{multirow}
\usepackage{subcaption}

\usepackage{tablefootnote}

\usepackage{pgfplotstable} %tikz plots data
\usepackage{pgfplots}
\pgfplotsset{compat=1.14}

\title{Pay less attention\\ with Lightweight and Dynamic Convolutions}
\author{Felix Wu\thanks{Work done during an internship at Facebook.} \\
Cornell University\\
\And
Angela Fan, Alexei Baevski, Yann N. Dauphin, Michael Auli \\
Facebook AI Research \\
}

\iclrfinalcopy % Uncomment for camera-ready version, but NOT for submission.
\begin{document}

\maketitle

\begin{abstract}
Self-attention is a useful mechanism to build generative models for language and images. 
It determines the importance of context elements by comparing each element to the current time step.
In this paper, we show that a very lightweight convolution can perform competitively to the best reported self-attention results.
Next, we introduce dynamic convolutions which are simpler and more efficient than self-attention.
We predict separate convolution kernels based solely on the current time-step in order to determine the importance of context elements. 
The number of operations required by this approach scales linearly in the input length, whereas self-attention is quadratic.
Experiments on large-scale machine translation, language modeling and abstractive summarization show that dynamic convolutions improve over strong self-attention models. 
On the WMT'14 English-German test set dynamic convolutions achieve a new state of the art of 29.7 BLEU.\footnote{Code and pre-trained models available at \url{http://github.com/pytorch/fairseq}} 
\end{abstract}

\section{Introduction}
There has been much recent progress in sequence modeling through recurrent neural networks (RNN; \citealt{sutskever2014sequence,bahdanau2015neural,wu2016google}), convolutional networks (CNN; \citealt{kalchbrenner2016nmt,gehring2016convolutional,gehring2017convs2s,kaiser2017depthwise}) and self-attention models \citep{paulus17intra,vaswani2017transformer}. 
RNNs integrate context information by updating a hidden state at every time-step, CNNs summarize a fixed size context through multiple layers, while as self-attention directly summarizes all context.

\begin{figure}[h]
\centering
\begin{subfigure}[b]{0.45\textwidth}
\centering
\includegraphics[width=1\textwidth]{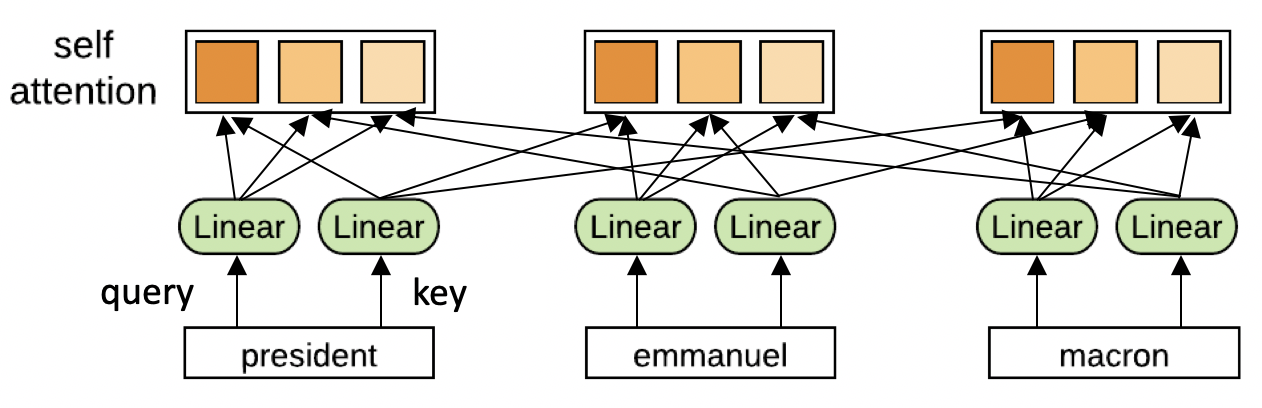}
\caption{Self-attention}
\label{fig:sattn2}
\end{subfigure}
\qquad \quad
\begin{subfigure}[b]{0.45\textwidth}
\centering
\includegraphics[width=1\textwidth]{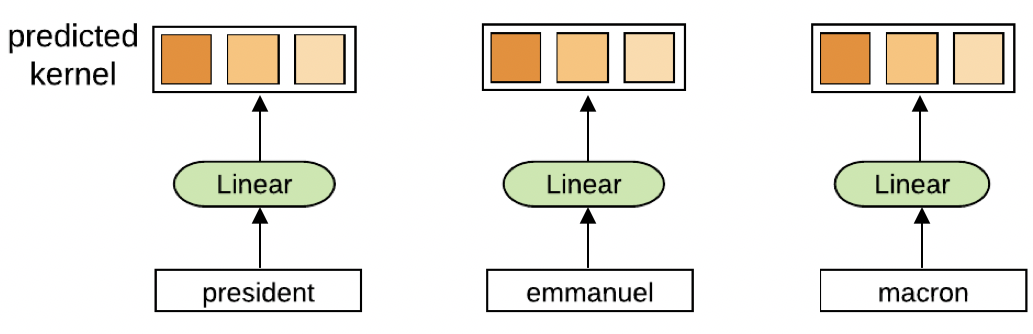}
\caption{Dynamic convolution}
\label{fig:dsdconv}
\end{subfigure}
\caption{Self-attention computes attention weights by comparing all pairs of elements to each other (\subref{fig:sattn2}) while as dynamic convolutions predict separate kernels for each time-step (\subref{fig:dsdconv}).}
\label{fig:attention}
\end{figure}

Attention assigns context elements \emph{attention weights} which define a weighted sum over context representations \citep{bahdanau2015neural,sukhbaatar2015memnet,chorowski2015asr,luong2015effective}.
Source-target attention summarizes information from another sequence such as in machine translation while as self-attention operates over the current sequence.
Self-attention has been formulated as \emph{content-based} where attention weights are computed by comparing the current time-step to all elements in the context (Figure~\ref{fig:sattn2}). 
The ability to compute comparisons over such unrestricted context sizes are seen as a key characteristic of self-attention \citep{vaswani2017transformer}.

However, the ability of self-attention to model long-range dependencies has recently come into question \citep{tang2018why} and the unlimited context size is computationally very challenging due to the quadratic complexity in the input length. 
Furthermore, in practice long sequences require the introduction of hierarchies \citep{liu2018wikisum}. 

In this paper, we introduce \emph{\sdconvfull{}s} which are depth-wise separable \citep{sifre2014rigid,Chollet2017XceptionDL,kaiser2017depthwise}, softmax-normalized and share weights over the channel dimension.
The result is a convolution with several orders of magnitude fewer weights than a standard non-separable convolution. 
Different to self-attention, \sdconvfull{}s reuse the same weights for context elements, regardless of the current time-step.

\emph{\Tvsdconvfull{}s} build on \sdconvfull{}s by predicting a different convolution kernel at every time-step. The kernel is a function of the current time-step only as opposed to the entire context as in self-attention (Figure~\ref{fig:dsdconv}).
\Tvsdconvfull{}s are similar to locally connected layers in the sense that the weights change at every position, however, the difference  is that weights are dynamically generated by the model rather than fixed after training \citep{lecun1998grad,taigman2014deepface,chen2015interspeech}.
Our approach also bears similarity to location-based attention which does not access the context to determine attention weights, however, we do not directly take the attention weights from the previous time-step into account \citep{chorowski2015asr,luong2015effective}.
\citet{shen2018block} reduce complexity by performing attention within blocks of the input sequence and 
\citet{shen2017disan,shen2018fast} perform more fine-grained attention over each feature.
\citet{shen2018emnlp} and \citet{gong2018emnlp} use input-dependent filters for text classification tasks.

Our experiments show that \sdconvfull{}s perform competitively to strong self-attention results and that \tvsdconvfull{}s can perform even better.
On WMT English-German translation \tvsdconvfull{}s achieve a new state of the art of 29.7 BLEU, on WMT English-French they match the best reported result in the literature, and on IWSLT German-English \tvsdconvfull{}s outperform self-attention by 0.8 BLEU.
\Tvsdconvfull{}s achieve 20\% faster runtime than a highly-optimized self-attention baseline.
For language modeling on the \gbw{} benchmark \tvsdconvfull{}s perform as well as or better than self-attention and on \cnndm{} abstractive document summarization we outperform a strong self-attention model.

\section{Background}\label{sec:background}
We first outline sequence to sequence learning and self-attention. Our work builds on non-separable convolutions as well as depthwise separable convolutions.

\paragraph{Sequence to sequence learning } maps a source sequence to a target sequence via two separate networks such as in machine translation \citep{sutskever2014sequence}.
The encoder network computes representations for the source sequence such as an English sentence and the decoder network autoregressively generates a target sequence based on the encoder output.

\paragraph{The self-attention} module of \citet{vaswani2017transformer} applies three projections to the input $\emX \in \R^{n \times d}$ to obtain key (K), query (Q), and value (V) representations, where $n$ is the number of time steps, $d$ the input/output dimension (Figure~\ref{fig:sattn}). 
It also defines a number of heads $H$ where each head can learn separate attention weights over $d_k$ features and attend to different positions.
The module computes dot-products between key/query pairs, scales to stabilize training, and then softmax normalizes the result. 
Finally, it computes a weighted sum using the output of the value projection (V):
\[
\mathrm{Attention}(Q,K,V) = \mathrm{softmax}(\frac{QK^T}{\sqrt{d_k}})V
\]

\paragraph{Depthwise convolutions} perform a convolution independently over every channel. The number of parameters can be reduced from $d^2k$ to $dk$ where $k$ is the kernel width.
The output $\emO \in \R^{n \times d}$ of a depthwise convolution with weight $\emW \in \R^{d \times k}$ for element $i$ and output dimension $c$ is defined as:
\[
\emO_{i, c} = \text{DepthwiseConv}(X, \emW_{c,:}, i, c) = \sum_{j=1}^{k} \emW_{c, j} \cdot \emX_{(i + j - \ceil{\frac{k+1}{2}}), c} 
\]

% %% to show the font of each symbol
% $\ra \rva \erva$\\
% $\rmA \ermA \vzero \va$ \\
% $\eva \mA \tA \gA $\\
% $\sA \emA \etA $

\section{Lightweight Convolutions}\label{sec:sdconv}
\begin{figure}[t]
\centering
\begin{subfigure}[b]{0.35\textwidth}
\centering
\includegraphics[width=1\textwidth]{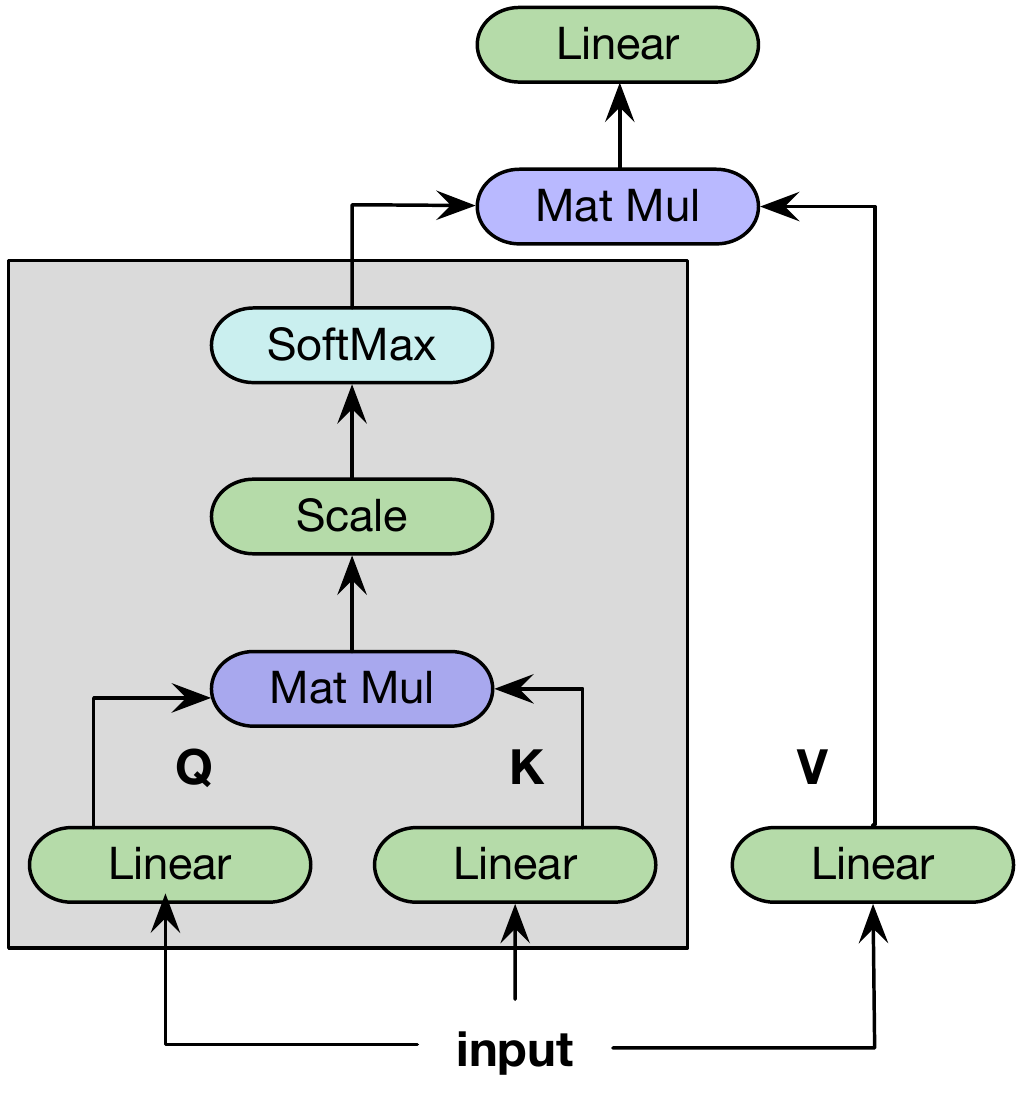}
\caption{Self-attention}
\label{fig:sattn}
\end{subfigure}
~ %add desired spacing between images, e. g. ~, \quad, \qquad, \hfill etc. 
% \qquad
\begin{subfigure}[b]{0.3\textwidth}
\centering
\includegraphics[width=0.34\textwidth]{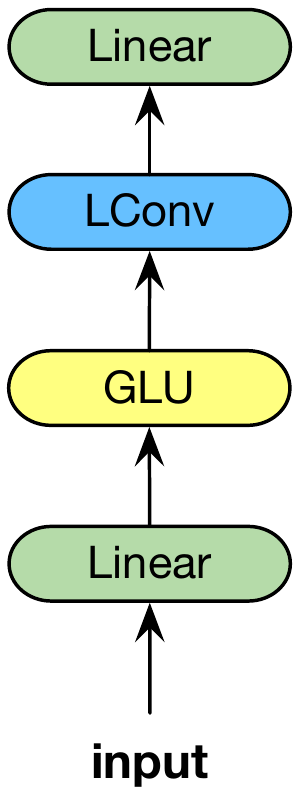}
\caption{\Sdconvfull{}}
\label{fig:sdconv}
\end{subfigure}
~ %add desired spacing between images, e. g. ~, \quad, \qquad, \hfill etc. 
% \qquad
\begin{subfigure}[b]{0.3\textwidth}
\centering
\includegraphics[width=0.86\textwidth]{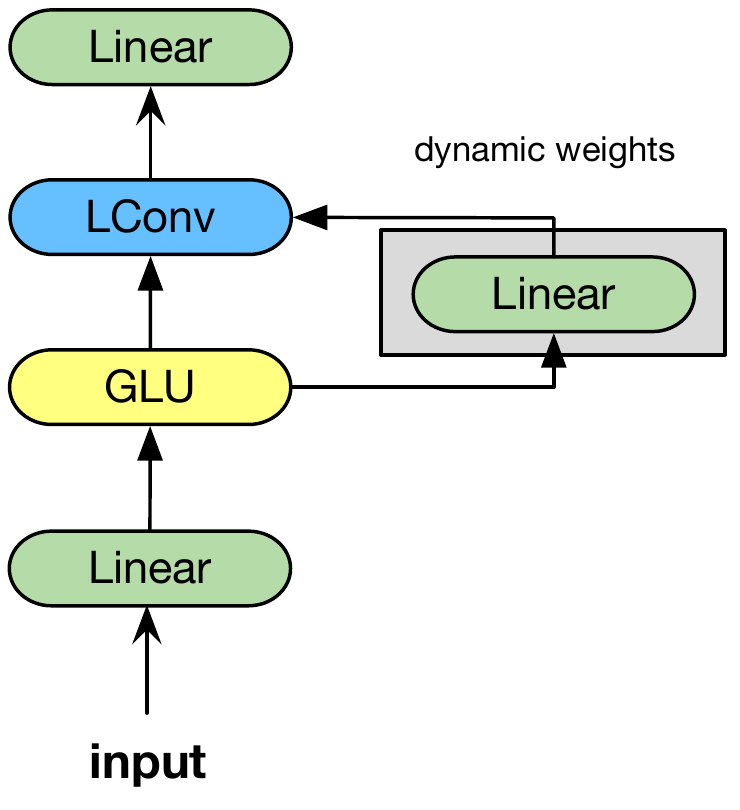}
\caption{Dynamic convolution}
\label{fig:tvsdconv}
\end{subfigure}
\caption{Illustration of self-attention, \sdconvfull{}s and \tvsdconvfull{}s.}
\label{fig:architectures}
\end{figure}

In this section, we introduce \newterm{\sdconv{}}, a depthwise convolution which shares certain output channels and whose weights are normalized across the temporal dimension using a softmax. 
Compared to self-attention, \sdconv{} has a fixed context window and it determines the importance of context elements with a set of weights that do not change over time steps.
We will show that models equipped with \sdconvfull{}s show better generalization compared to regular convolutions and that they can be competitive to state-of-the-art self-attention models (\textsection\ref{sec:results}). 
This is surprising because the common belief is that content-based self-attention mechanisms are necessary to obtaining state-of-the-art results in natural language processing applications.
Furthermore, the low computational profile of \sdconv{} enables us to formulate efficient \tvsdconvfull{}s (\textsection\ref{sec:tvsdconv}).

\sdconv{} computes the following for the $i$-th element in the sequence and output channel $c$:
\[
\text{\sdconv{}}(X, \emW_{\ceil{\frac{cH}{d}},:}, i, c) = \text{DepthwiseConv}(X, \text{softmax}(\emW_{\ceil{\frac{cH}{d}},:}), i, c)
\]

\paragraph{Weight sharing.} 
We tie the parameters of every subsequent number of $\frac{d}{H}$ channels, which reduces the number of parameters by a factor of $\frac{d}{H}$. 
As illustration, a regular convolution requires 7,340,032 ($d^2 \times k$) weights for $d=1024$ and $k=7$, a depthwise separable convolution has 7,168 weights ($d \times k$), and with weight sharing, $H=16$, we have only 112 ($H \times k$) weights. 
We will see that this vast reduction in the number of parameters is crucial to make dynamic convolutions possible on current hardware.
\citet{wang2018smoothed} ties the weights of all channels (H = 1).

\paragraph{Softmax-normalization.}
We normalize the weights $\emW \in \R^{H \times k}$ across the temporal dimension $k$ using a softmax operation:
\[
\text{softmax}(\emW)_{h,j} = \frac{\exp{\emW_{h, j}}}{\sum_{j'=1}^k \exp{\emW_{h, j'}}}
\]

\paragraph{Module.} Figure~\ref{fig:sdconv} shows the architecture of the module where we integrate \sdconv{}. 
We first apply an input projection mapping from dimension $d$ to $2d$, followed by a gated linear unit (GLU;  \citealt{dauphin2017convlm}), and the actual \sdconvfull{}.
The GLU uses half of the inputs as gates by applying sigmoid units and then computes a pointwise product with the other inputs.
We also apply an output projection of size $\emW^O \in \R^{d \times d}$ to the output of \sdconv{}.

\paragraph{Regularization. }
We found DropConnect to be a good regularizer for the \sdconv{} module \citep{wan2013regularization}.
Specifically, we drop every entry of the normalized weights $softmax(\emW)$ with probability $p$ and divide it by $1-p$ during training. 
This amounts to removing some of the temporal information within a channel.

\paragraph{Implementation.}
Existing CUDA primitives for convolutions did not perform very well to implement \sdconv{} and we found the following solution faster on short sequences:
We copy and expand the normalized weights $\emW \in \mathbb{R}^{H \times k}$ to a band matrix of size $BH \times n \times n$, where $B$ is the batch size.
We then reshape and transpose the inputs to size $BH \times n \times \frac{d}{H}$, and perform a batch matrix multiplication to get the outputs.
We expect a dedicated CUDA kernel to be much more efficient.

\section{Dynamic convolutions}\label{sec:tvsdconv}

A dynamic convolution has kernels that vary over time as a learned function of the individual time steps.
A dynamic version of standard convolutions would be impractical for current GPUs due to their large memory requirements.
We address this problem by building on \sdconv{} which drastically reduces the number of parameters (\textsection\ref{sec:sdconv}). 

\newterm{\tvsdconv{}} takes the same form as \sdconv{} but uses a time-step dependent kernel that is computed using a function $f:\mathbb{R}^{d}\to \R^{H \times k}$:
\[
\text{\tvsdconv{}}(X, i, c) = \text{\sdconv{}}(X, f(\emX_i)_{h,:}, i, c)
\]
we model $f$ with a simple linear module with learned weights $\emW^Q \in \R^{H \times k \times d}$, i.e.,
$ f(\emX_i) = \sum_{c=1}^d \emW^Q_{h, j, c} \emX_{i, c}$. 

Similar to self-attention, \tvsdconv{} changes the weights assigned to context elements over time.
However, the weights of \tvsdconv{} do not depend on the entire context, they are a function of the current time-step only.
Self-attention requires a quadratic number of operations in the sentence length to compute attention weights, while the computation of dynamic kernels for \tvsdconv{} scales linearly in the sequence length. 

Our experiments (\textsection\ref{sec:results}) show that models using \tvsdconv{} match or exceed the performance of state-of-the-art models that use context-based self-attention. 
This challenges the typical intuitions about the importance of content-based self-attention in natural language processing applications.

\section{Experimental setup}\label{sec:setup}

\subsection{Model Architecture}\label{sec:architecture}

We use an encoder-decoder architecture for sequence to sequence learning \citep{sutskever2014sequence} and we closely follow the architectural choices presented in \citet{vaswani2017transformer}.
Our self-attention baseline is the fairseq re-implementation of the Transformer Big architecture  \citep{ott2018scaling}.\footnote{\url{https://github.com/pytorch/fairseq}}

The encoder and decoder networks have $N$ blocks each.
Encoder blocks contain two sub-blocks: The first is a self-attention module (\textsection\ref{sec:background}), a \sdconv{} module (\ref{sec:sdconv}), or a \tvsdconv{} module (\textsection\ref{sec:tvsdconv}).
The second sub-block is a feed-forward module: $ReLU(\emW^1 \emX + b_1) \emW^2 + b_2$ where $\emW^1 \in \R^{d \times d_{ff}}$, $\emW^2 \in \R^{d_{ff} \times d}$ and $d=1024$, $d_{ff}=4096$ unless otherwise stated.
Sub-blocks are surrounded by residual connections \citep{he2015deep} and layer normalization \citep{ba2016layer}. 

Decoder blocks are identical except that they have an additional source-target attention sub-block between the self-attention and feed-forward module. 
The source-target attention is equivalent to the self-attention module, except that the values and keys are projections over the encoder output for each source word.

Words are fed to the encoder and decoder networks in $d$ dimensional embeddings. We add sinusoidal position embeddings to encode the absolute position of each word in the sequence \citep{kaiser2017depthwise,vaswani2017transformer}.
The model computes a distribution over vocabulary $V$ by transforming the decoder output via a linear layer with weights $\emW^V \in \R^{d \times V}$ followed by softmax normalization. 

\sdconv{} and \tvsdconv{} are identical to Transformer Big, except that self-attention modules are swapped with either fixed or dynamic convolutions.
These models also use fewer parameters per block (cf. Figure~\ref{fig:sdconv} and Figure~\ref{fig:tvsdconv}) and we 
therefore increase the number of blocks to $N=7$ for the encoder to roughly match the parameter count of Transformer Big.
We generally set $H=16$.
Both \sdconv{} and \tvsdconv{} set the the encoder and decoder kernel sizes to 3, 7, 15, 31x4 for each block respectively; except for the decoder where we have only three top layers with kernel size 31.

\subsection{Datasets and Evaluation}\label{sec:datasets}

To get a thorough understanding of the limitations of \sdconv{} and \tvsdconv{} we evaluate on three different tasks: machine translation, language modeling and abstractive summarization.

\noindent {\bf Machine Translation.}
We report results on four benchmarks: 
For WMT English to German (En-De) we replicate the setup of~\citet{vaswani2017transformer}, based on WMT'16 training data with 4.5M sentence pairs, we validate on newstest2013 and test on newstest2014.\footnote{\scriptsize \url{https://github.com/tensorflow/tensor2tensor/blob/321bacaa3abcca5dbf341ed6fb3d4a1531e513ff/tensor2tensor/data_generators/translate_ende.py\#L60-L63}}
The vocabulary is a 32K joint source and target byte pair encoding (BPE; \citealt{sennrich2016bpe}).
For WMT English to French (En-Fr), we borrow the setup of~\citet{gehring2017convs2s} with 36M training sentence pairs from WMT'14, validate on newstest2012+2013 and test on newstest2014.
The 40K vocabulary is based on a joint source and target BPE factorization.

For WMT English to Chinese (Zh-En), we pre-process the WMT'17 training data following \citet{hassan2018parity} resulting in 20M sentence pairs. 
We develop on devtest2017 and test on newstest2017. 
For IWSLT'14 German-English (De-En) we replicate the setup of \citet{edunov2018classical} for 160K training sentence pairs and 10K joint BPE vocabulary.
For this benchmark only, data is lowercased.

For \ende{}, \enfr{}, we measure case-sensitive tokenized BLEU.\footnote{\scriptsize{\url{https://github.com/moses-smt/mosesdecoder/blob/master/scripts/generic/multi-bleu.perl}}}
For \ende{} only we apply compound splitting similar to~\citet{vaswani2017transformer}.
For \zhen{} we measure detokenized BLEU to be comparable to \citet{hassan2018parity}.\footnote{SacreBLEU hash: \texttt{\scriptsize
BLEU+case.mixed+lang.zh-en+numrefs.1+smooth.exp+test.wmt17+tok.13a+version.1.2.11}}

We train three random initializations of a each configuration and report test accuracy of the seed which resulted in the highest validation BLEU. 
Ablations are conducted on the validation set and we report the mean BLEU and standard deviation on this set.
\ende{}, \enfr{} are based on beam search with a beam width of 5, \iwslt{} uses beam 4, and \zhen{} beam 8 following \citet{hassan2018parity}.
For all datasets, we tune a length penalty as well as the number of checkpoints to average on the validation set.

\noindent {\bf Language Modeling.} 
We evaluate on the large-scale \gbw{} dataset \citep{chelba2013one} which contains 768M tokens and has a vocabulary of nearly 800K types. 
Sentences in this dataset are shuffled and we batch sentences independently of each other.
Models are evaluated in terms of perplexity on the valid and test portions.

\noindent {\bf Summarization.}
We test the model's ability to process long documents on the \cnndm{} summarization task \citep{hermann15,nallapati16} comprising over 280K news articles paired with multi-sentence summaries. 
Articles are truncated to 400 tokens \citep{see17} and we use a BPE vocabulary of 30K types \citep{fan17}. 
We evaluate in terms of F1-Rouge, that is Rouge-1, Rouge-2 and Rouge-L \citep{lin04}.\footnote{We use the following parameters for \texttt{ROUGE-1.5.5.pl}: -m -a -n 2}
When generating summaries, we follow standard practice in tuning the maximum output length, disallowing repeating the same trigram, and we apply a stepwise length penalty \citep{paulus17intra,fan17,wu2016google}.

\subsection{Training and Hyperparameters}

\noindent {\bf Translation.}
We use a dropout rate of 0.3 for WMT En-De and IWSLT De-En, 0.1 for WMT En-Fr, and 0.25 for WMT Zh-En. 
WMT models are optimized with Adam and a cosine learning rate schedule \citep{kingma2015adam,loshchilov2016cosine} where the learning rate is first linearly warmed up for 10K steps from $10^{-7}$ to $10^{-3}$ and then annealed following a cosine rate with a single cycle. 
For IWSLT'14 De-En, we use a schedule based on the inverse square root of the current step \citep{vaswani2017transformer}.
We train the WMT models on 8 NVIDIA V100 GPUs for a total of 30K steps on WMT En-De, 40K steps for WMT Zh-En and 80K steps for WMT En-Fr. 
For IWSLT De-En we train for 50K steps on a single GPU.

We use floating point 16 precision and accumulate the gradients for 16 batches before applying an update \citep{ott2018scaling}, except for IWSLT where we do not accumulate gradients.
Batches contain up to 459K source tokens and the same number of target tokens for both WMT En-De and WMT Zh-En, 655K for En-Fr, and 4K for IWSLT De-En.
We use label smoothing with $0.1$ weight for the uniform prior distribution over the vocabulary  \citep{szegedy2015inception,pereyra2017regularize}.

\noindent {\bf Language Modeling.} 
We follow the same setup as for translation but remove the encoder module.
For the \gbw{} benchmark we use an adaptive softmax output layer to reduce the computational burden of the large vocabulary \citep{grave2016arxiv,press2017using} and tie it with variable sized input word embeddings (Anonymous et al., 2018). 
The first 60K types in the adaptive softmax have dimension $1024$, the 100K types dimension $256$, and the last 633K types have size $64$.

We train on 32 GPUs with batches of 65K tokens for 975K updates. As optimizer we use Nesterov's accelerated gradient method \citep{sutskever2013icml} with a momentum value of $0.99$ and we re-normalize gradients if their norm exceeds $0.1$ \citep{pascanu2013difficulty}.
The learning rate is linearly warmed up from $10^{-7}$ to $1$ for 16K steps and then annealed using a cosine learning rate schedule \citep{loshchilov2016cosine} with one cycle.

\noindent {\bf Summarization.}
We train with Adam using the cosine learning rate schedule with a warmup of 10K steps and a period of 20K updates. 
We use weight decay 1e-3 and dropout 0.3. %, and train with label smoothing 0.1. 

\section{Results}\label{sec:results}
\subsection{Machine Translation}\label{sec:translation}

We first report results on \ende{} and \enfr{} where we compare to the best results in the literature, most of which are based on self-attention.
Table~\ref{tab:mt1} shows that \sdconv{} performs very competitively and only trails the state of the art result by 0.1 BLEU on \enfr{}; the state of the art is based on self-attention \citep{ott2018scaling}.
This is despite the simplicity of \sdconv{} which operates with a very small number of fixed weights over all time steps whereas self-attention computes dot-products with all context elements at every time-step.

\tvsdconv{} outperforms the best known result on \ende{} by 0.4 BLEU and achieves a new state of the art, whereas on \enfr{} it matches the state of the art.
This shows that content-based self-attention is not necessary to achieve good accuracy on large translation benchmarks.

\begin{table*}[t]
\centering
\begin{tabular}{lcrr}
\toprule
Model & Param (En-De) & \ende{} & \enfr{} \\
\midrule
\citet{gehring2017convs2s} & 216M & 25.2 & 40.5 \\
\citet{vaswani2017transformer} & 213M & 28.4 & 41.0 \\
\citet{ahmed2017WeightedTN} & 213M & 28.9 & 41.4 \\
\citet{chen2018arxiv} & 379M & 28.5 & 41.0 \\
\citet{shaw2018relpos}  & - & 29.2 & 41.5 \\
\citet{ott2018scaling} & 210M & 29.3 & \textbf{43.2} \\
\midrule
\sdconv{} & 202M & 28.9 & 43.1 \\
\tvsdconv{} & 213M & \textbf{29.7} & \textbf{43.2} \\
\bottomrule
\end{tabular}
\caption{Machine translation accuracy in terms of BLEU for \ende{} and \enfr{} on newstest2014.}
\label{tab:mt1}
\end{table*}

\iwslt{} is a much smaller benchmark and we therefore switch to a smaller architecture: $d_{ff} = 1024$, $d=512$, and $H=4$. 
The self-attention baseline on this dataset is the best reported result in the literature (Table~\ref{tab:mt2}).\footnote{We omit comparison to \citet{elbayad2018pervasiveattn} since their test set is not directly comparable.}
\sdconv{} outperforms this baseline by 0.4 BLEU and \tvsdconv{} improves by 0.8 BLEU.
We further run experiments on \zhen{} translation to evaluate on a non-European language.
\sdconv{} outperforms the baseline by 0.5 BLEU and \tvsdconv{} by 0.6 BLEU. 

\begin{table*}[t]
\centering
\begin{tabular}{lcrrr}
\toprule
Model & Param (Zh-En) & \iwslt{} & \zhen{} \\
\midrule
\citet{deng2018latent} & - & 33.1 & - \\
\cite{hassan2018parity} & - & - & 24.2 \\
\midrule
Self-attention baseline & 292M & 34.4 & 23.8 \\
\sdconv{} & 285M & 34.8 & 24.3 \\
\tvsdconv{} & 296M & \textbf{35.2} & \textbf{24.4} \\
\bottomrule
\end{tabular}
\caption{Machine translation accuracy in terms of BLEU on \iwslt{} and \zhen{}.}
\label{tab:mt2}
\end{table*}

\subsection{Model ablation}\label{sec:ablation}

In this section we evaluate the impact of the various choices we made for \sdconv{} (\textsection\ref{sec:sdconv}) and \tvsdconv{} (\textsection\ref{sec:tvsdconv}).
We first show that limiting the maximum context size of self-attention has no impact on validation accuracy (Table~\ref{tab:ablation}).
Note that our baseline is stronger than the original result of \citet{vaswani2017transformer}.
Next, we replace self-attention blocks with non-separable convolutions (CNN) with kernel size 3 and input/output dimension $d=1024$. 
The CNN block has no input and output projections compared to the baseline and we add one more encoder layer to assimilate the parameter count. 
This CNN with a narrow kernel trails self-attention by 1 BLEU.

We improve this result by switching to a depthwise separable convolution (CNN Depthwise) with input and output projections of size $d=1024$. 
When we progressively increase the kernel width from lower to higher layers then this further improves accuracy. 
This narrows the gap to self-attention to only 0.5 BLEU.
DropConnect gives a slight performance improvement and weight sharing does not decrease performance.
Adding softmax normalization to the weights is only 0.3 BLEU below the accuracy of the baseline. 
This corresponds to \sdconv{}.
In Appendix~\ref{app:smnorm} we compare softmax-normalization to various alternatives.
Finally, \tvsdconvfull{}s (\tvsdconv{}) achieve the same validation accuracy as self-attention with slightly fewer parameters and at 20\% higher inference speed.
Softmax-normalization is important for \tvsdconv{} since training diverged in our experiments when removing it.
To make the models more comparable, we do not introduce GLU after the input projection.

For comparison, we re-implemented averaged attention networks (AAN; \citealt{zhang2018aan}) which compute a uniform average over past model states instead of a weighted average as in self-attention.
Our re-implementation is efficient: we measure 129 sentences/sec for a base transformer-AAN on newstest2014 compared to 20 sentences/sec for \citet{zhang2018aan}.
Table~\ref{tab:ablation} shows that our models outperform this approach.
Note that AANs still use self-attention in the encoder network while as our approach does away with self-attention both in the encoder and decoder.

\begin{table*}[t]
\centering
\begin{tabular}{lcrr}
\toprule
Model & Param & BLEU & Sent/sec \\
\midrule
\citet{vaswani2017transformer} & 213M & 26.4 & - \\
Self-attention baseline (k=$\inf$, H=16) & 210M & 26.9 $\pm$ 0.1 & 52.1 $\pm$ 0.1 \\
Self-attention baseline (k=3,7,15,31x3, H=16) & 210M & 26.9 $\pm$ 0.3 & 54.9 $\pm$ 0.2 \\
\midrule
CNN (k=3) & 208M & 25.9 $\pm$ 0.2 & 68.1 $\pm$ 0.3 \\
CNN Depthwise (k=3, H=1024) & 195M & 26.1 $\pm$ 0.2 & 67.1 $\pm$ 1.0 \\
+ Increasing kernel (k=3,7,15,31x4, H=1024) & 195M  & 26.4 $\pm$ 0.2 & 63.3 $\pm$ 0.1 \\
+ DropConnect (H=1024) & 195M & 26.5 $\pm$ 0.2 & 63.3 $\pm$ 0.1 \\
+ Weight sharing (H=16) & 195M & 26.5 $\pm$ 0.1 & 63.7 $\pm$ 0.4 \\
+ Softmax-normalized weights [\sdconv] (H=16) & 195M & 26.6 $\pm$ 0.2 & 63.6 $\pm$ 0.1 \\
+ Dynamic weights [\tvsdconv] (H=16) & 200M & 26.9 $\pm$ 0.2 & 62.6 $\pm$ 0.4 \\
Note: \tvsdconv (H=16) w/o softmax-normalization & 200M & diverges  \\
\midrule
AAN decoder + self-attn encoder & 260M & 26.8 $\pm$ 0.1 & 59.5 $\pm$ 0.1 \\
AAN decoder + AAN encoder & 310M & 22.5 $\pm$ 0.1 & 59.2 $\pm$ 2.1 \\
\bottomrule
\end{tabular}
\caption{Ablation on WMT English-German newstest2013. 
(+) indicates that a result includes \emph{all} preceding features. 
Speed results based on beam size 4, batch size 256 on an NVIDIA P100 GPU.}
\label{tab:ablation}
\end{table*}

\subsection{Language Modeling}\label{sec:lm}

As second task we consider language modeling on the \gbw{} benchmark. 
The self-attention baseline has $N=16$ blocks, each with a self-attention module and a feed-forward module using $d_{ff} = 4096$ and $d=1024$.
\tvsdconv{} uses $N=17$ blocks to assimilate the parameter count and we use kernel sizes 15x2, 31x4 and 63x11.
Table~\ref{tab:gbw_lm} shows that \tvsdconv{} achieves slightly better perplexity than our self-attention baseline which is very competitive.

\begin{table*}
\centering
\begin{tabular}{lcrrr}
\toprule
Model & Param & Valid & Test \\
\midrule
2-layer LSTM-8192-1024 \citep{jozefowicz2016lm} & -- & -- & 30.6 \\
Gated Convolutional Model \citep{dauphin2017convlm} & 428M & -- & 31.9  \\
Mixture of Experts \citep{shazeer2017outrageously} & 4371M $^\dagger$ & -- & 28.0 \\
\midrule
Self-attention baseline & 331M & 26.67 & 26.73 \\  
\tvsdconv{} & 339M & 26.60 & \textbf{26.67} \\ 
\bottomrule
\end{tabular}
\caption{Language modeling results on the Google Billion Word test set. \\
\small{$^\dagger$does not include embedding and softmax layers}}
\label{tab:gbw_lm}
\end{table*}

\subsection{Abstractive Summarization}\label{sec:abs}

Finally, we evaluate on the \cnndm{} abstractive document summarization benchmark where we encode a document of up to 400 words and generate multi-sentence summaries. 
This tests the ability of our model to deal with longer sequences.
We reduce model capacity by setting $d=1024$, $d_{ff}=2048$, $H=8$, similar to the Transformer base setup of \citet{vaswani2017transformer}.

Table~\ref{tab:abs} shows that \sdconv{} outperforms the self-attention baseline as well as comparable previous work and \tvsdconv{} performs even better.
We also show results for a reinforcement learning approach \citep{celikyilmaz2018deep} and note that RL is equally applicable to our architecture.\footnote{An earlier version of this paper erroneously compared to \citet{gehrmann2018bottom}, however, their setup is based on the full-text \cnndm{} whereas we use the more common entity-anonymized version.}

\begin{table*}[t]
\centering
\begin{tabular}{lcrrr}
\toprule
Model & Param & Rouge-1 & Rouge-2 & Rouge-l \\    
\midrule
LSTM \citep{paulus17intra} & - & 38.30 & 14.81 &  35.49  \\ 
CNN \citep{fan17} & - & 39.06 & 15.38 & 35.77 \\
\midrule
Self-attention baseline & 90M & 39.26 & 15.98 & 36.35  \\
\sdconv & 86M & 39.52 & 15.97 & 36.51  \\
\tvsdconv & 87M & \textbf{39.84} & \textbf{16.25} & \textbf{36.73} \\
\midrule 
RL \citep{celikyilmaz2018deep} & - & \textbf{41.69} & \textbf{19.47} & \textbf{37.92} \\ 
\bottomrule
\end{tabular}
\caption{Results on \cnndm{} summarization. We compare to likelihood trained approaches except for \citet{celikyilmaz2018deep}.}
\label{tab:abs}
\end{table*}

\section{Conclusion}\label{sec:conclusion}

We presented \sdconvfull{}s which perform competitively to the best reported results in the literature despite their simplicity. 
They have a very small parameter footprint and the kernel does not change over time-steps. 
This demonstrates that self-attention is not critical to achieve good accuracy on the language tasks we considered.

\Tvsdconvfull{}s build on \sdconvfull{}s by predicting a different kernel at every time-step, similar to the attention weights computed by self-attention. 
The dynamic weights are a function of the current time-step only rather than the entire context.

Our experiments show that \sdconvfull{}s can outperform a strong self-attention baseline on WMT'17 Chinese-English translation, \iwslt{}'14 German-English translation and \cnndm{} summarization. 
\Tvsdconvfull{}s improve further and achieve a new state of the art on the test set of WMT'14 English-German.
Both \sdconvfull{} and \tvsdconvfull{} are 20\% faster at runtime than self-attention.
On \gbw{} language modeling we achieve comparable results to self-attention.

We are excited about the future of dynamic convolutions and plan to apply them to other tasks such as question answering and computer vision where inputs are even larger than the tasks we considered in this paper.


\bibliography{master}
\bibliographystyle{iclr2019_conference}

\clearpage
\appendix
\section*{Supplementary Material}

\section{Comparison of softmax-normalization to alternatives}
\label{app:smnorm}

We compare our proposed softmax-normalization of weights to other alternatives in Table~\ref{tab:ablation_softmax}. 
For each setting, we use three seeds and report the mean and the standard deviation of the BLEU score on WMT English-German newstest2013. 
The softmax and norms are computed over the kernel dimension. 
Simply using the absolute value of the weights or squaring them does not make the training more stable, which shows that having all non-negative weights is not critical. 
Dividing the weights by the $\ell_2$-norm or bounding the weights with sigmoid or the hyperbolic tangent function also stablizes the training procedure; however, the softmax-normalization performs best.

\begin{table*}[h]
\centering
\setlength\extrarowheight{5pt}
\begin{tabular}{lr}
\toprule
Method & BLEU \\
\midrule
$\emW$ (No normalization) &  diverges \\
$\softmax (\emW)$ & 26.9 $\pm$ 0.2 \\
$\sigmoid (\emW)$ & 26.6 $\pm$ 0.3 \\
$\tanh (\emW)$ & 25.6 $\pm$ 0.2 \\
$\frac{\emW}{\lVert \emW \rVert_1 + \epsilon}$ & diverges \\
$\frac{\emW}{\lVert \emW \rVert_2 + \epsilon}$ & 26.8 $\pm$ 0.2 \\
$ \text{power} (\emW, 2) $ & diverges \\
$\text{abs} (\emW)$ & diverges \\
$\frac{\text{abs} (\emW)}{\lVert \emW \rVert_1 + \epsilon}$ & diverges \\
$\frac{\text{abs} (\emW)}{\lVert \emW \rVert_2 + \epsilon}$ & 26.7 $\pm$ 0.2 \\
\bottomrule
\end{tabular}
\caption{Alternatives to softmax-normalization in \tvsdconv{} on WMT English-German newstest2013 ($\epsilon= 10^{-6}$).}
\label{tab:ablation_softmax}
\end{table*}

\section{On the current state of non-autoregressive generation}

\begin{table*}[h]
\small
\centering
\resizebox{\textwidth}{!}{  
\begin{tabular}{ l c r r r}
\toprule
Model (batch size = 1, beam size = 1) & Param & BLEU & Sent/sec & Tok/sec \\
\midrule
NAT (+ FT) \citep{gu2018nonautoregressive} & - & 17.7 & 25.6 & - \\
NAT (+ FT + NPD=10) \citep{gu2018nonautoregressive} & - & 18.7 & 12.7 & - \\
NAT (+ FT + NPD=100) \citep{gu2018nonautoregressive} & - & 19.2 & 3.9 & - \\
LT, Improved Semhash \citep{kaiser2018fastdecoding} &- & 19.8 & 9.5 & - \\
\midrule
IR $i_{dec} = 1$ \citep{lee2018deterministic} & - & 13.9 & - & 511.4 \\
IR $i_{dec} = 2$ \citep{lee2018deterministic} & - & 17.0 & - & 393.6 \\
IR $i_{dec} = 5$ \citep{lee2018deterministic} & - & 20.3 & - & 139.7 \\
IR $i_{dec} = 10$ \citep{lee2018deterministic} & - & 21.6 & - & 90.4 \\
IR Adaptive \citep{lee2018deterministic} & - & 21.5 & - & 107.2 \\
\midrule
NART w/ hints \citep{li2019hintbased} & - & 21.1 & 38.5 & - \\
NART w/ hints ($B=4$, 9 candidates) \citep{li2019hintbased} & - & 25.2 & 22.7 & - \\
\midrule
ENAT Embedding Mapping \citep{guo2018non} & - & 20.7 & 41.7 & - \\
\multirow{2}{*}{\shortstack[l]{ENAT Embedding Mapping (rescoring 9 candidates) \\ \citep{guo2018non}}} & \multirow{2}{*}{-} & \multirow{2}{*}{24.3} & \multirow{2}{*}{20.4} &  \multirow{2}{*}{-} \\
\\
\midrule
Autoregressive \citep{gu2018nonautoregressive} & - & 22.7 & 2.5 & - \\ 
Autoregressive \citep{lee2018deterministic} & - & 23.8 & - & 54.0 \\ 
Transformer \citep{li2019hintbased} & - & 27.3 & 1.3 & - \\ 
Transformer \citep{guo2018non} & - & 27.4 & 1.6 & - \\ 
\midrule
\tvsdconv{} (1-decoder layer (k=31)) & 124M & 26.1 & 15.2 & 423.0   \\
\tvsdconv{} (3-decoder layers (k=3,7,15)) & 153M & 27.7 & 7.2 & 202.3   \\
\tvsdconv{} (6-decoder layers (k=3,7,15,31,31,31)) & 200M & 28.5 & 3.9 & 110.9  \\
\bottomrule
\end{tabular}
}
\caption{Inference speed of non-autoregressive models and small decoder versions of \tvsdconv{} on WMT English-German newstest2014. For some models, the decoding speed (sent/sec) is derived by taking the inverse of the sentence generation latency in the literature.}
\label{tab:smalldec}
\end{table*}

In this section we compare \tvsdconv{} to current non-autoregressive models in the literature. 
We measured generation speed for \tvsdconv{} on a P100 GPU using batch size one to be comparable with other results.
Results in the literature are based on either NVIDIA GTX-1080 GPUs or P100 GPUs. 
The effects of different GPU types is likely negligible because GPUs are vastly underutilized with batch size one.

Table~\ref{tab:smalldec} shows that \tvsdconv{} with a single decoder layer outperforms all previously reported non-autoregressive results both in terms of speed as well as accuracy.
Only two non-autoregressive concurrent efforts \citep{guo2018non,li2019hintbased} achieve a speedup over \tvsdconv{} with a small drop in BLEU.
Notably, both \citet{guo2018non} and \citet{li2019hintbased} distill autoregressive models into non-autoregressive models \citep{hinton2015distilling}, in order to improve their results.
% Thus, a larger gap of BLEU score may exist if the auto-regressive models are also trained with the knowledge distillation method.

\end{document}